# Korean-English Machine Translation with Multiple Tokenization Strategy

**Dojun Park, Youngjin Jang** and **Harksoo Kim**
Konkuk University Natural Language Processing Lab.
`dojun.parkk@gmail.com, {danyon,nlpdrkim}@konkuk.ac.kr`

## Abstract

This work was conducted to find out how tokenization methods affect the training results of machine translation models. In this work, alphabet tokenization, morpheme tokenization, and BPE tokenization were applied to Korean as the source language and English as the target language respectively, and the comparison experiment was conducted by repeating 50,000 epochs of each 9 models using the Transformer neural network. As a result of measuring the BLEU scores of the experimental models, the model that applied BPE tokenization to Korean and morpheme tokenization to English recorded 35.73, showing the best performance.

## 1 Introduction

Tokenization refers to the operation of segmenting input sentences into tokens, which are appropriate units according to specific criteria, and each token formed by the tokenizing operation is converted into a word vector, forming a numerical meaning that can be interpreted by computers. Therefore, the consideration of what criteria to construct tokens for input sentences before performing the machine translation task is as important as the design of the neural network structure.

The tokenization should be considered differently depending on which languages are dealt with. This is because the types of characters and the method of notation are different for each language. Korean and English to be covered in this work have different writing systems. Korean uses Hangul, which consists of 24 basic alphabets, and English uses 26 Latin alphabets and is distinguished between uppercase and lowercase letters. In addition, Hangul follows the syllable-based writing rule in which the letters constituting one syllable are gathered to form an independent letter, but in English, each letter is arranged in horizontal order, so separate syllable segments are not specified.

In this work, we applied three tokenization methods to reflect the unique literal and linguistic characteristics of Korean and English in machine translation models. The first one is alphabet tokenization, which separates individual consonants and vowels into individual tokens, the second one is morpheme tokenization, which separates words into morphemes, which are the smallest units that form linguistic meanings, and the third one is Byte Pair Encoding (BPE) tokenization, which learns and separates character pairs that frequently appear together as new tokens. By applying the above three tokenization methods to Korean and English data, the individual model will train and the performance differences of each model represented by BLEU scores will be measured.

## 2 Related Work

In [1], Korean-English and English-Korean machine translation models were trained by applying a shared attention model, and tokenization of words, syllables, and phonemes was applied to Korean sentences, and BPE and character tokenization were applied to English sentences. As a result of measuring the BLEU score of the trained models, in both Korean-English and English-Korean translation model training, the model that applied word tokenization to Korean and BPE tokenization to English showed the best performance. In [2], an English-Korean machine translation model using the LSTM model applying attention was built, and BPE and WPM tokenization were applied to Korean and English respectively. As a result of the experiment, the model that applied BPE tokenization to both English and Korean showed the highest BLEU score. In [3], a study was conducted on Korean-English and English-Korean machine translation models using transformer architecture, and experimental models were trained by applying phoneme, syllable, morpheme, subword, and morpheme-aware subword tokenization to Korean. morpheme-aware subword tokenization is a method of applying BPE tokens after dividing original sentences into morpheme units. In the experimental result, the model applied morpheme-aware subword tokenization to Korean showed the best performance in both Korean-English and English-Korean translations.

## 3 Suggested Model

| Tokenization | Language | Tokenized sentences |
|---|---|---|
| Alphabet Tokenization | Korean | ㅇ/ㅏ/ㄴ/ㄴ/ㅕ/ㅇ/ㅎ/ㅏ/ㅅ/ㅔ/ㅇ/ㅛ/. |
| | English | n / i / c / e / æ / t / o / æ / m / e / e / t / æ / y / o / u / . |
| Morpheme Tokenization | Korean | 안녕 / æ / 하세요 / . |
| | English | nice / æ / to / æ / meet / æ / you / . |
| BPE Tokenization | Korean | 안녕 / 하 / 세@@ / 요 / . |
| | English | ni@@ / ce / to / me@@ / et / you / . |

Table 1: Examples of tokenization methods for Korean and English sentences

This work tokenized Korean as the source language and English as the target language as shown in Table 1. First of all, in order to reflect the spacing rules of Korean and English before the token separation in alphabet tokenization and morpheme tokenization, spaces were replaced with the special character ''æ'' (U+00E6). For alphabet tokenization to Korean, the hgtk library was used to separate the consonants and vowels combined according to the syllable-based writing rule, and in the case of English alphabet tokenization, an additional library was not used because the individual alphabets were used as individual tokens. Okt morpheme analyzer of koNLPy[4] was used for Korean morpheme tokenization, and English morpheme analyzer of spaCy was used for English tokenization. In the case of BPE tokenization[5], the character sets that appear simultaneously with high frequency were learned using the built-in module of the OpenNMT-py[6] used to build transformer models, and based on this, BPE tokenization for the input sentences was performed. In BPE tokenization, a token is created by adding "@@" (U+0040) after the token following the next token without spaces so that spaces can be distinguished. After that, in the model evaluation process, all special characters inserted for spacing rules were removed through post-processing.

The number of tokens formed after tokenization is as follows. In the case of alphabet tokenization, 81 Korean and 50 English tokens were created, and in the case of morpheme tokenization, 692,838 Korean and 161,817 English tokens each, and in the case of BPE tokenization, 37,015 Korean and 32,485 English tokens were generated.

## 4 Experiment

### 4.1 Experimental data and methods

In this work, we used the Korean-English parallel news corpus consisting of 800,000 sentence pairs provided by AI Hub[1)] as data for model training. The reason for using the news corpus was that the stylistic consistency, refined vocabulary and sentences used in news articles met the conditions suitable for training a translation model. The Korean-English news corpus, consisting of 800,000 sentences in total, is randomly selected at a ratio of 98:1:1 and classified into training data (784,000 sentences), validation data (1,000 sentences), and evaluation data (1,000 sentences). Model training, model validation, and model evaluation were performed using this divided data.

For the experiment, Transformer[7] neural network with identical hyperparameters was constructed using the OpenNMT-py, and the hyperparameters applied to the model designing are as follows. The number of layers of the encoder and decoder is 6, the output dimension of the encoder and decoder is 512, the number of layers of the feed-forward neural network is 2048, the head of the multi-head attention is 8, the drop out is 0.1, and the batch size was unified to 4096 so that model training was performed under the same conditions.

### 4.2 Result

#### 4.2.1 Analysis of BLEU Scores

Table 2 shows the results of measuring BLEU[8] score of the models after saving the training models at every 10,000 epochs during the training process in which each model repeats 50,000 epochs. BLEU scores in Table 2 showed a low score of less than 1 in most of the models, indicating that model training did not proceed properly.

|  |  | Epoch | EN | | |
|---|---|---|---|---|---|
|  |  |  | Alphabet | Morpheme | BPE |
| KO | Alphabet | 10,000 | 0.00 | **0.12** | 0.03 |
|  |  | 20,000 | 0.00 | 0.11 | 0.03 |
|  |  | 30,000 | 0.03 | 0.09 | 0.00 |
|  |  | 40,000 | **0.04** | 0.08 | **0.05** |
|  |  | 50,000 | 0.03 | 0.07 | 0.00 |
|  | Mor-pheme | 10,000 | 0.00 | **0.37** | 0.28 |
|  |  | 20,000 | 0.00 | 0.18 | 0.19 |
|  |  | 30,000 | 0.00 | 0.10 | 0.30 |
|  |  | 40,000 | 0.00 | 0.10 | 0.13 |
|  |  | 50,000 | **0.07** | 0.08 | **0.31** |
|  | BPE | 10,000 | 0.08 | 30.28 | 17.73 |
|  |  | 20,000 | **0.09** | 34.50 | 21.84 |
|  |  | 30,000 | 0.08 | 35.36 | 22.78 |
|  |  | 40,000 | 0.08 | 35.70 | 23.22 |
|  |  | 50,000 | 0.08 | **35.73** | **23.54** |

Table 2: BLEU score of each model measured every 10,000 epochs

On the other hand, the model in which BPE tokenization is applied to both Korean and English shows a BLEU score between 17 and 24. It was 17.73 in 10,000 epochs and 21.84 in 20,000 epochs, followed by a slight upward trend, recording 23.54 in 50,000 epochs.

The model that showed the highest BLEU score in this experiment was the model that applied BPE tokenization to Korean and morpheme tokenization to English. It shows 30.28 in 10,000 epochs, which is a higher score exceeding the BLEU score of 50,000 epochs of the model that applied BPE tokenization to both Korean and English. After showing 34.50 at 20,000 epochs, it continued a slight upward tendency, and recorded 35.73 at 50,000 epochs, recording the highest BLEU score in this experiment.

Prior studies [1] and [2] differ from this study in that they used the OpenSubtitles Korean-English parallel corpus composed of relatively short sentences as training data. Besides, since the measurement of the performance of the models of the previous studies and the models of this study are limited to the comparison of BLEU scores, there is a clear difficulty in generalizing the superiority and inferiority of the objective performance. However, under the experimental conditions of this study, it can be seen that the strategy of applying BPE tokenization to Korean and morpheme tokenization to English was the most effective tokenization method.

## 5 Conclusion

In this work, a total of 9 models were trained with the strategy to apply alphabet, morpheme, and BPE tokenization to Korean and English respectively. As a result of measuring BLEU scores of the models, BLEU scores of less than 1 were measured in the remaining 7 models excluding the Korean BPE-English morpheme tokenization model and the Korean BPE-English BPE tokenization model.

The Korean BPE-English morpheme tokenization model showed the best performance with a BLEU score of 35.73, followed by the Korean BPE-English BPE tokenization model with a BLEU score of 23.54.

However, in that BLEU simply evaluates the output sentences based on the recall of the correct answer sentences, there is a limitation in evaluating the completeness of the result when considering the characteristics of natural language that can be expressed in synonyms and other sentence structures. Therefore, we plan to conduct a follow-up study on the intrinsic evaluation method that can measure the performance of machine translation models based on linguistic features of the output sentences of machine translation models.